\documentclass[Numbered]{sn-jnl}

\usepackage{graphicx}%
\usepackage{multirow}%
\usepackage{amsmath,amssymb,amsfonts}%
\usepackage{amsthm}%
\usepackage{mathrsfs}%
\usepackage[title]{appendix}%
\usepackage{xcolor}%
\usepackage{textcomp}%
\usepackage{manyfoot}%
\usepackage{booktabs}%
\usepackage{algorithm}%
\usepackage{algorithmicx}%
\usepackage{algpseudocode}%
\usepackage{listings}%

\newtheorem{theorem}{Theorem}
\newtheorem{remark}{Remark}%
\newtheorem{assumption}{Assumption}

\newtheorem{definition}{Definition}%

\DeclareMathOperator*{\argmax}{argmax}

\newcommand{\R}{\mathbb{R}} 
\newcommand{\N}{\mathbb{N}}
\newcommand{\bs}{\boldsymbol}

\usepackage{bbm}

\begin{document}

\title[A comprehensive theoretical framework for the optimization of neural networks classification performance with respect to weighted metrics]{A comprehensive theoretical framework for the optimization of neural networks classification performance with respect to weighted metrics}


\author*[1]{\fnm{Francesco} \sur{Marchetti}}\email{francesco.marchetti@math.unipd.it}

\author[2]{\fnm{Sabrina} \sur{Guastavino}}\email{guastavino@dima.unige.it}

\author[2]{\fnm{Cristina} \sur{Campi}}\email{campi@dima.unige.it}

\author[2]{\fnm{Federico} \sur{Benvenuto}}\email{benvenuto@dima.unige.it}

\author[2]{\fnm{Michele} \sur{Piana}}\email{piana@dima.unige.it}

\affil*[1]{\orgdiv{Department of Mathematics \lq\lq Tullio Levi-Civita\rq\rq}, \orgname{University of Padova}, \orgaddress{\street{Via Trieste 63}, \city{Padova}, \postcode{35121}, \country{Italy}}}

\affil[2]{\orgdiv{Department of Mathematics DIMA}, \orgname{University of Genova}, \orgaddress{\street{Via Dodecaneso 33}, \city{Genova}, \postcode{16146}, \country{Italy}}}


\abstract{In many contexts, customized and weighted classification scores are designed in order to evaluate the goodness of the predictions carried out by neural networks. However, there exists a discrepancy between the maximization of such scores and the minimization of the loss function in the training phase. In this paper, we provide a complete theoretical setting that formalizes weighted classification metrics and then allows the construction of losses that drive the model to optimize these metrics of interest. After a detailed theoretical analysis, we show that our framework includes as particular instances well-established approaches such as classical cost-sensitive learning, weighted cross entropy loss functions and value-weighted skill scores.}

\keywords{Weighted classification metrics, score oriented losses, value-weighted scores, deep learning}



\maketitle

\section{Introduction}

\subsection{The supervised classification problem via neural networks}

Neural networks are in the state-of-art of many machine learning classification tasks in a huge variety of contexts \cite{Goodfellow16}, such as, e.g., medical diagnostics \cite{kermany2018identifying}, forecasting problems \cite{guastavino2022prediction,guastavino2022implementation}, image classification \cite{Pelletier19}. Thanks to their flexibility, they have been adapted and shaped to be employed in the solution of different complex issues. In this paper we focus on their usage in addressing the classical supervised learning problem, which can be described by using the following notation. Let $\mathcal{X}\subset \Omega,\; \Omega\subset\mathbb{R}^m$ be a training set of data in some space of dimension $m\in\mathbb{N}_{\ge 1}$, and let $\mathcal{Y}$ be a finite set of $d$ \textit{labels} or \textit{classes} to be learned, where $d\in\mathbb{N}_{\ge 2}$. In \textit{multiclass} tasks, each element in $\mathcal{X}$ is uniquely assigned to a class in $\mathcal{Y}$, and it is common for these labels to be integer or \textit{one-hot} encoded \cite{Harris13}. Differently, in the so-called \textit{multilabel} framework, each element in $\mathcal{X}$ can be assigned to more than one class in $\mathcal{Y}$.

Hence, the supervised classification problem consists in constructing a function $\hat{y}$ on $\Omega$ by \textit{learning} from the labeled data set $\mathcal{X}$, so that $\hat{y}$ \textit{models} the data-label relation between elements in $\Omega$ and labels in $\mathcal{Y}$. In the context of neural networks, such function is characterized by a vector (or matrix) of weights $\bs{\theta}$, i.e., $\hat{y}=\hat{y}_{\bs{\theta}}(\bs{x})$, $\bs{x}\in\Omega$. These weights are set by minimizing a certain loss function $\ell(\hat{y}_{\bs{\theta}}(\bs{x}),y)$, that measures the possible discrepancy between the prediction given for $\bs{x}$ and its true label $y\in\mathcal{Y}$. More precisely, the minimization process is carried out on the training set, that is we consider the task
\begin{equation}\label{eq:obj_fun}
    \min_{\bs{\theta}}\ell(\hat{y}_{\bs{\theta}}(\bs{x}),y),\; \bs{x}\in\mathcal{X},
\end{equation}
where  possible regularization terms on the weights of the network can be also added. Beyond a good performance on the training set, the model is expected to generalize in predicting unseen test samples in $\Omega$. Although several loss functions have been proposed, with the aim of accounting for the specific properties of the problem under analysis \cite{Janocha16,Liu16,Zhu20}, the most common choice as loss function is the Cross Entropy (CE) and its generalizations \cite{Good52,Zhang18}, which is provided with a robust theoretical background that originated in the context of information theory. 

\subsection{Classical and weighted scores}

After the neural network model weights are calibrated in the training phase, the classification results are evaluated by means of metrics, also known as \textit{scores}, that are chosen according to the framework and are usually built from the elements of the so-called \textit{confusion matrix} (CM), which provides a consistent description of the quality and the type of the errors achieved by the constructed classifier. In this work, we mainly focus on the cardinal binary classification setting $d=2$ where the CM can be expressed in its fundamental form that involves a \textit{positive} $\{y=1\}$ and a \textit{negative} $\{y=0\}$ class, and the output of the network is such that $\hat{y}_{\bs{\theta}}\in (0,1)$, accordingly.
The choice of the evaluation metric is crucial for the assessment of the predictions: for example, in cases where the dataset is imbalanced, the ratio between the correct predictions over the number of elements, that is the \textit{accuracy}, is not very meaningful, while other scores like the F$_1$ score, True Skill Statistic (TSS), or the Heidke Skill Score (HSS) are more appropriate for evaluating the goodness of predictions. 
Another perspective to evaluate the predictions consists in assigning a different impact to errors of different type: this falls into the cost-sensitive learning field \cite{elkan2001foundations}. In this framework, the evaluation of the predicted value is
commonly carried out on the basis of preassigned costs for \textit{False Positives} (FPs)
and \textit{False Negatives} (FNs). For example, in applications as medical diagnosis, or identification of frauds, missing a positive class is worse than incorrectly classifying an example from the negative class: therefore a major cost to FNs is assigned with respect to FPs by defining a suitable cost matrix \cite{fernandez2018cost,thai2010cost}.
Furthermore, in cases where binary predictions are performed over time, such scores defined upon the classical CM do not take into account the distribution of the predictions over the actual outcomes: 
a false positive is counted as \lq\lq one\rq\rq$\;$independently it represents a false alarm just before an actual occurrence or a false alarm given after an actual occurrence. 
The evaluation of the forecast in terms of its usefulness
to support the user while making a decision is known in literature as the forecast value. In \cite{guastavino2021}, value-weighted skill scores have been introduced in order to take into account the severity of errors with respect to the distribution of predictions over time, with applications ranging from weather, space weather forecasting, through environment \cite{guastavino2022operational,hu2022probabilistic}.

\subsection{Addressing the score-loss discrepancy}

Loss minimization and score maximization are intertwined concepts. However, a direct maximization of a skill score of interest in the training phase is not recommended, as the score is usually discontinuous with respect to the weights of the network. This is due to the fact that the \textit{continuous} predictions in the interval $(0,1)$ are assigned to the negative or to the positive class by assessing their relative value with respected to a fixed threshold $\tau\in (0,1)$. In many situations, this threshold parameter is set to $\tau=0.5$, but it can be tuned \textit{a posteriori} in order to maximize a chosen skill score.
In order to deal with this discrepancy between loss minimization and score maximization, in literature many empirical strategies have been designed to align the loss function to approximate metrics of interest \cite{Huang19,Rezatofighi19,Singh10}. In particular, in \cite{MARCHETTI2022108913} a new class of Score-Oriented Loss (SOL) functions has been introduced. To build such losses, the threshold $\tau$ is treated not as a fixed value, but as a random variable provided with a certain \textit{a priori} density function. Then, considering the expected value of the entries of the CM, it is possible to obtain an \textit{averaged} score that is derivable with respect to the weights of the model. During the training process, the averaged score maximization automatically leads to classifiers that are oriented to achieve results that are already optimal for the chosen score, being the optimal threshold value also driven by the \textit{a priori} probability density function. We point out that the cost-sensitive learning approach proposed in \cite{pmlr-v37-narasimhana15,narasimhan2021training} shares a similar spirit with \cite{MARCHETTI2022108913}, meaning that their common objective is optimizing the classification process with respect to weighted losses. However, on the one hand, in \cite{narasimhan2021training} the classifier is obtained by taking the argmax of the outputs of the network, possibly in a multiclass case, without considering thresholds that convert probability outcomes in 0-1 classification predictions. On the other hand, \cite{MARCHETTI2022108913} focuses on the influence of the threshold value in a threshold-based classification process, and it is then suitable for a generalization to the multilabel case, as we will deepen in this paper.

\subsection{Outline and contribution of this work}

In this work, our first purpose is to provide a wide theoretical background for the formalization of weighted classification scores. This is done in Section \ref{sec:conf_to_score}, where starting from classical CMs we then derive weighted CMs and scores. Then, in Section \ref{sec:weighted_matrices} we analyze the presented weighted matrices via a probabilistic approach, which is a crucial step that allows the definition of weighted SOL (wSOL) functions in Section \ref{sec:wsol}, which are designed to valorize the chosen weighted metrics directly in the training phase of the network. Therefore, we show the effectiveness of the proposed theoretical framework by analyzing some particular cases that are represented in literature: classical cost-sensitive learning approaches in Section \ref{sec:cost}, weighted cross entropy loss functions in Section \ref{sec:ce} and value-weighted skill scores in Section \ref{sec:valuew}. The extension of the achieved results to the multilabel setting is outlined in Section \ref{sec:multilabel}. Finally, we draw some final remarks in Section \ref{sec:conclusions}.

\section{From confusion matrices to (weighted) classification scores}\label{sec:conf_to_score}

To facilitate our analysis, we define $\mathcal{S}=\mathcal{S}_n(\bs{\theta})=\{(\hat{y}_{\bs{\theta}}(\bs{x}_i),y_i)\}_{i=1,\dots,n}$ to be a batch of predictions-labels, in which $y_i\in\{0,1\}$ is the true label associated to the element $\bs{x}_i\in\mathcal{X}$. Letting $\tau\in\R$ be a threshold parameter, $\tau\in(0,1)$, the classical confusion matrix is defined as
\begin{equation*}
\mathrm{CM}(\tau,\mathcal{S}_n(\bs{\theta})) = 
\begin{pmatrix}
\mathrm{TN}(\tau,\mathcal{S}_n(\bs{\theta})) & \mathrm{FP}(\tau,\mathcal{S}_n(\bs{\theta})) \\
\mathrm{FN}(\tau,\mathcal{S}_n(\bs{\theta})) & \mathrm{TP}(\tau,\mathcal{S}_n(\bs{\theta}))
\end{pmatrix},
\end{equation*}
being
\begin{equation}\label{eq:mat_defi}
    \begin{split}
         & \mathrm{TN}(\tau,\mathcal{S}_n(\bs{\theta})) = \sum_{i=1}^n{(1-y_i)\mathbbm{1}_{\{\hat{y}_{\bs{\theta}}(\bs{x}_i)<\tau\}}}, \quad \mathrm{TP}(\tau,\mathcal{S}_n(\bs{\theta})) = \sum_{i=1}^n{y_i\mathbbm{1}_{\{\hat{y}_{\bs{\theta}}(\bs{x}_i)>\tau\}}},\\
         & \mathrm{FP}(\tau,\mathcal{S}_n(\bs{\theta})) = \sum_{i=1}^n{(1-y_i)\mathbbm{1}_{\{\hat{y}_{\bs{\theta}}(\bs{x}_i)>\tau\}}},\quad \mathrm{FN}(\tau,\mathcal{S}_n(\bs{\theta})) = \sum_{i=1}^n{y_i\mathbbm{1}_{\{\hat{y}_{\bs{\theta}}(\bs{x}_i)<\tau\}}},
        \end{split}
\end{equation}
defined in terms of the indicator function
\begin{equation*}
    \mathbbm{1}_{\{x>\tau\}}=\begin{cases} 0 & \textrm{if } x\le \tau,\\
        1 & \textrm{if }x>\tau,\end{cases}\quad x\in(0,1).
\end{equation*}
\begin{definition}\label{def:score}
    A score (or classification metric) $s:M_{2,2}(\N) \longrightarrow \R$ is a function that takes in input the elements of $\mathrm{CM}(\tau,\mathcal{S}_n(\bs{\theta}))$ and gives a real number as output, which is non-decreasing with respect to $\mathrm{TN}(\tau,\mathcal{S}_n(\bs{\theta}))$ and $\mathrm{TP}(\tau,\mathcal{S}_n(\bs{\theta}))$ and non-increasing with respect to $\mathrm{FN}(\tau,\mathcal{S}_n(\bs{\theta}))$ and $\mathrm{FP}(\tau,\mathcal{S}_n(\bs{\theta}))$.
\end{definition}
In the following, our purpose is to extend the presented framework to include modified scores where false positives and negatives are weighted. Proceeding in this way, we remark that we recover the classical CM in case of perfect classification, as we do not modify the true positives and negatives. Therefore, a positive-valued weight function $w$ is introduced in the FP and FN entrances of the CM, obtaining
\begin{equation}\label{eq:weighted_general}
\begin{split}
   & \mathrm{wFP}=
     \sum_{i=1}^n{w(1-y_i)\mathbbm{1}_{\{\hat{y}_{\bs{\theta}}(\bs{x}_i)>\tau\}}}, \quad \mathrm{wFN}=
    \sum_{i=1}^n{wy_i\mathbbm{1}_{\{\hat{y}_{\bs{\theta}}(\bs{x}_i)<\tau\}}}.
\end{split}
\end{equation}
In general, this weight function $w$ can depend on different inputs. We consider the following ones.
\begin{itemize}
    \item 
    The label $y_i$. The weight can treat differently the case $y_i=0$ from $y_i=1$, i.e., treat false negatives and false positives in different manners. For example, this turns out to be significant in medical diagnosis applications where false negatives are more \textit{dangerous} than false positives.
    \item
    The prediction $\hat{y}_{\bs{\theta}}(\bs{x}_i)$, and the threshold $\tau$. Predictions could be handled in regards to their relative value with respect to the threshold. As an example, fixed $\tau=0.4$, a false positive $\hat{y}_{\bs{\theta}}(\bs{x}_i)=0.5$ can be treated as a minor mistake compared to $\hat{y}_{\bs{\theta}}(\bs{x}_i)=0.9$.
    \item
    The batch $\mathcal{S}_n(\bs{\theta})$. Indeed, this is the case where the CM can be provided with a \textit{probabilistic} connotation by normalizing the entrances \cite{Koco13}.
    \item
    Past and future labels and predictions, in case the data are chronological. Letting $T\in\mathbb{N}$, we can consider as inputs of the weight function the vectors
    \begin{equation}\label{eq:passato_futuro_pre}
    \begin{split}
        &\bs{y}_i= (y_{i-T},\dots,y_{i+T}), \quad \bs{\mathbbm{1}}_i= \big(\mathbbm{1}_{\{\hat{y}_{\bs{\theta}}(\bs{x}_{i-T})>\tau\}},\dots,\mathbbm{1}_{\{\hat{y}_{\bs{\theta}}(\bs{x}_{i+T})>\tau\}}\big).
    \end{split}
    \end{equation}
    Note that while in \eqref{eq:weighted_general} the elements of the batch $\mathcal{S}_n(\bs{\theta})$ are not required to be in chronological order, we assume a chronological order in \eqref{eq:passato_futuro_pre} meaning that we are looking at the \textit{time} interval $[i-T,i+T]$ centered at time $i$. This little abuse allows us to keep a simpler notation. This framework is particularly suitable to formalize situations in which positive predictions are interpreted as \textit{alarms}, and then, for example, a missed alarm (false negative) may be almost negligible if alarms were produced by close past samples. We will deepen this setting in Section \ref{sec:valuew}.
\end{itemize}
Therefore, the weight function may assume the form
\begin{equation*}
    w=w\big(\tau,y_i,\hat{y}_{\bs{\theta}}(\bs{x}_i),\mathcal{S}_n(\bs{\theta}),\bs{y}_i,\bs{\mathbbm{1}}_i\big).
\end{equation*}
We denote as $\mathrm{wCM}(\tau,\mathcal{S}_n(\bs{\theta}))$ the weighted confusion matrix where we replace FP and FN with wFP and wFN.
\begin{definition}\label{def:wscore}
    A weighted score $s_w:M_{2,2}(\N) \longrightarrow \R$ is a score $s$ that takes in input the weighted matrix $\mathrm{wCM}(\tau,\mathcal{S}_n(\bs{\theta}))$, i.e., $s_w=s(\mathrm{wCM}(\tau,\mathcal{S}_n(\bs{\theta})))$.
\end{definition}

\section{Expected confusion matrices}\label{sec:weighted_matrices}

In the following, we consider the non-weighted case $\mathrm{CM}(\tau,\mathcal{S}_n(\bs{\theta}))$ as a special case of $\mathrm{wCM}(\tau,\mathcal{S}_n(\bs{\theta}))$ where the weight function is set to $w\equiv 1$.

Assume that a certain weighted score $s_w$ has been designed in order to assess the performance of a classifier. As discussed in the introductory section, $s_w$ is discontinuous with respect to the weights of the neural network, and can not be directly used in constructing a loss function. Surely, we can not expect regularity with respect to the threshold $\tau$ and the indicator function. To deal with this problem, in what follows we leverage the approach effectively carried out in \cite{MARCHETTI2022108913} with classical CMs.

The first crucial step is to let $\tau$ be a continuous random variable whose probability density function (pdf) $f$ is supported in $[a,b]\subseteq[0,1]$, $a,b\in\R$. We denote as $F$ the cumulative density function (cdf)
\begin{equation*}
    F(x)=\int_{a}^{x}{f(\xi)\mathrm{d}\xi},\quad x\le b.
\end{equation*}
This ensures the possibility of averaging the entrances of  $\mathrm{wCM}$ with respect to the threshold, thus replacing the irregular indicator function with the \textit{regular} cdf $F$. Indeed, we remind that the expected value of the indicator function behaves as
\begin{equation*}
    \mathbb{E}_{\tau}[\mathbbm{1}_{\{x>\tau\}}]=\int_{a}^{b}{\mathbbm{1}_{\{x>\xi\}}f(\xi)\mathrm{d}\xi}=\int_{a}^{x}{f(\xi)\mathrm{d}\xi}=F(x).
\end{equation*}
Hence, we consider the expected value of the matrix $\mathrm{wCM}(\tau)=\mathrm{wCM}(\tau,\mathcal{S}_n(\bs{\theta}))$ with respect to $\tau$, meaning
\begin{equation*}
    \mathbb{E}_{\tau}[\mathrm{wCM}(\tau)]= \begin{pmatrix}
\mathbb{E}_{\tau}[\mathrm{TN}(\tau)] & \mathbb{E}_{\tau}[\mathrm{wFP}(\tau)] \\
\mathbb{E}_{\tau}[\mathrm{wFN}(\tau)] & \mathbb{E}_{\tau}[\mathrm{TP}(\tau)]
\end{pmatrix}.
\end{equation*}
As far as the true positives and negatives are concerned, we get
\begin{equation*}
        \begin{split}
         & \mathbb{E}_{\tau}[\mathrm{TN}(\tau)] = \sum_{i=1}^n{(1-y_i)(1-F(\hat{y}_{\bs{\theta}}(\bs{x}_i)))}, \quad \mathbb{E}_{\tau}[\mathrm{TP}(\tau)] = \sum_{i=1}^n{y_iF(\hat{y}_{\bs{\theta}}(\bs{x}_i))},
        \end{split}
\end{equation*}
where we used the linearity of the expected value. Then, in the assumptions of the previous section, we obtain
\begin{equation*}
\begin{split}
     & \mathbb{E}_{\tau}[\mathrm{wFP}(\tau)]=
     \sum_{i=1}^n{W_{\textrm{P}}(1-y_i)},\quad \mathbb{E}_{\tau}[\mathrm{wFN}(\tau)]=
    \sum_{i=1}^n{W_{\textrm{N}}\:y_i},
\end{split}
\end{equation*}
with 
{\small{
\begin{equation}\label{eq:pesos}
\begin{split}
   & W_{\textrm{P}}=
     \int_{a}^{b}{w\big(\xi,y_i,\hat{y}_{\bs{\theta}}(\bs{x}_i),\mathcal{S}_n(\bs{\theta}),\bs{y}_i,\bs{\mathbbm{1}}_i(\xi)\big)\mathbbm{1}_{\{\hat{y}_{\bs{\theta}}(\bs{x}_i)>\xi\}}f(\xi)\mathrm{d}\xi}, \\
   & W_{\textrm{N}}=
    \int_{a}^{b}{w\big(\xi,y_i,\hat{y}_{\bs{\theta}}(\bs{x}_i),\mathcal{S}_n(\bs{\theta}),\bs{y}_i,\bs{\mathbbm{1}}_i(\xi)\big)\mathbbm{1}_{\{\hat{y}_{\bs{\theta}}(\bs{x}_i)<\xi\}}f(\xi)\mathrm{d}\xi},
\end{split}
\end{equation}}}
where we highlighted the dependence of the indicator functions on the threshold. We conclude this section with the definition of admissible weight function and score.
\begin{definition}\label{def:admit}
    Let $w$ be a weight function. We say that $w$ is admissible if $W_{\textrm{P}}$ and $W_{\textrm{N}}$ are derivable with respect to the weights $\bs{\theta}$ of the neural network. Accordingly, we say that the weighted score $s_w$ is admissible if so is $w$.
\end{definition}

\section{Weighted score-oriented losses}\label{sec:wsol}

If $w$ is admissible according to Definition \ref{def:admit}, the classification metric can be finally employed in the construction of a score-oriented loss function.
\begin{definition}\label{def:scoloss}
Let $\mathcal{S}_n(\bs{\theta})$ be a batch of predictions-labels and
let $s_w$ be an admissible weighted score. We call a weighted SOL (wSOL) function related to $s_w$ the loss
\begin{equation*}
    \ell_{s_w}(\mathcal{S}_n(\bs{\theta})))= - s\big(\mathbb{E}_{\tau}[\mathrm{wCM}(\tau,\mathcal{S}_n(\bs{\theta}))]\big).
\end{equation*}
\end{definition}
With the following result, we highlight the benefit of minimizing the loss $\ell_{s_w}$ in the training phase.
\begin{theorem}\label{teor:cool}
    Recalling the optimization problem in \eqref{eq:obj_fun}, we have that
    \begin{equation}\label{eq:resultone}
    \min_{\bs{\theta}}\ell_{s_w}(\mathcal{S}(\bs{\theta}))\approx \max_{\bs{\theta}} \mathbb{E}_{\tau}[s(\mathrm{wCM}(\tau,\mathcal{S}_n(\bs{\theta})))],
    \end{equation}
    precisely:
    \begin{enumerate}
        \item 
        If the score $s$ is linear with respect to the entrances of the weighted CM, then equality is achieved in \eqref{eq:resultone}.
        \item
        If the score $s$ is analytic and non-linear with respect to the entrances of the weighted CM, then equality is achieved in \eqref{eq:resultone} up to derivative terms (see \eqref{eq:taylor}).
    \end{enumerate}
\end{theorem}
\begin{proof}
    If the score $s$ is linear with respect to the entrances of the weighted CM, we can exploit the linearity of the expected value and obtain
    \begin{equation*}
    \begin{split}
        \mathbb{E}_{\tau}[s(\mathrm{wCM}(\tau,\mathcal{S}_n(\bs{\theta})))]&=s(\mathbb{E}_{\tau}[\mathrm{wCM}(\tau,\mathcal{S}_n(\bs{\theta}))])=-\ell_{s_w}(\mathcal{S}_n(\bs{\theta}))),
    \end{split}
    \end{equation*}
    which implies
    \begin{equation*}
        \min_{\bs{\theta}}\ell_{s_w}(\mathcal{S}(\bs{\theta}))= \max_{\bs{\theta}} \mathbb{E}_{\tau}[s(\mathrm{wCM}(\tau,\mathcal{S}_n(\bs{\theta})))].
    \end{equation*}
    Now, assume $s$ is analytic with respect to the entrances of the weighted CM. Then, using the abridged notations $s_w=s(\mathrm{wCM}(\tau,\mathcal{S}_n(\bs{\theta})))$ and\\ $\overline{s_w}=s(\mathbb{E}_{\tau}[\mathrm{wCM}(\tau,\mathcal{S}_n(\bs{\theta}))])$, we can leverage the Taylor formula (see e.g. \cite{Benaroya05})
    \begin{equation}\label{eq:taylor}
           \mathbb{E}_{\tau}[s_w]= \overline{s_w}+\sum_{|\bs{\alpha}|\ge 1}{\frac{D^{\bs{\alpha}}\overline{s_w}}{\bs{\alpha}!}\mathbb{E}_{\tau}[(s_w-\overline{s_w})^{\bs{\alpha}}]},
    \end{equation}
    where the entrances of the weighted CM are \textit{vectorized} in $\mathbb{R}^4$ and $\bs{\alpha}\in\mathbb{N}^4$ denotes the classical multi-index notation for derivatives of multivariate functions.
\end{proof}

The result in Theorem \ref{teor:cool} certifies that the model is effectively driven to valorize the score of interest directly in the training phase. Furthermore, since the expected value is calculated with respect to the chosen pdf for $\tau$, the network is steered to optimize the score with respect to the threshold values that are taken into account by the pdf.

\begin{remark}
    The proposed framework can include a multi-objective setting where different scores are considered. Indeed, letting $s_w^1,\dots,s_w^m$ be $m\in\mathbb{N}$ weighted admissible scores and due to the linearity of the expected value, we may optimize the convex combination
    \begin{equation*}
    \begin{split}
         & s_w^{comb} = \beta_1s_w^1+\dots+\beta_ms_w^m,\\
         &\beta_1,\dots,\beta_m\ge0,\;\beta_1+\dots+\beta_m=1,   
    \end{split}
    \end{equation*}
    by considering
    \begin{equation*}
            \ell_{s_w^{comb}}= \beta_1\ell_{s_w^1}+\dots+\beta_m\ell_{s_w^m}.
    \end{equation*}
\end{remark}

\section{Applications to cost-sensitive learning}\label{sec:cost}

In this section, we show how the proposed setting includes as an instance classical cost-sensitive learning approaches \cite{elkan2001foundations}, where a cost matrix with non-negative entrances
\begin{equation*}
\mathrm{CC} = 
\begin{pmatrix}
\mathrm{C}_{00} & \mathrm{C}_{01} \\
\mathrm{C}_{10} & \mathrm{C}_{11}
\end{pmatrix}
\end{equation*}
is applied to the CM via the pointwise (or Hadamard) product $   \mathrm{CC}\odot \mathrm{CM}(\tau,\mathcal{S}_n(\bs{\theta}))
$. The matrix CC is employed to assign different costs to each classification outcome. In particular, in most applications the diagonal elements are set to zero values, i.e., $\mathrm{C}_{00}=\mathrm{C}_{11}=0$, or simply ignored \cite{thai2010cost,Zadrozny03}. The reason behind this choice is that the focus is put on weighting different errors in different manners, which turns out to be useful in cases where false negatives and false positives are not of equal importance, or when it is necessary to treat a strong imbalance in the dataset, as we outlined in the introduction. This approach is included in our theoretical framework by considering the admissible weight function
\begin{equation*}
    w_{cost}(y_i)=(1-y_i)\mathrm{C}_{01}+y_i\mathrm{C}_{10}.
\end{equation*}
Then, by setting $w=w_{cost}$ and computing the expected value with respect to $\tau$, we get
\begin{equation*}
\begin{split}
    & \mathbb{E}_{\tau}[\mathrm{wFP}(\tau)] = \mathrm{C}_{01}\sum_{i=1}^n{(1-y_i)F(\hat{y}_{\bs{\theta}}(\bs{x}_i))},\\
    & \mathbb{E}_{\tau}[\mathrm{wFN}(\tau)] = \mathrm{C}_{10}\sum_{i=1}^n{y_i(1-F(\hat{y}_{\bs{\theta}}(\bs{x}_i)))},\\    
\end{split}
\end{equation*}
as $w_{cost}$ is independent of the threshold $\tau$. Therefore, it is possible to consider a cost-sensitive score that relies on the weighted errors and derive the corresponding wSOL. As an example, we can consider the score
\begin{equation}\label{eq:scost}
    s^{cost}= -(\mathrm{FP}+\mathrm{FN}),
\end{equation}
and then the loss
\begin{equation*}
    \ell_{s_w^{cost}}(\mathcal{S}_n(\bs{\theta}))= \mathbb{E}_{\tau}[\mathrm{wFP}(\tau,\mathcal{S}_n(\bs{\theta}))]+\mathbb{E}_{\tau}[\mathrm{wFN}(\tau,\mathcal{S}_n(\bs{\theta}))],
\end{equation*}
where we highlighted the dependence on $\mathcal{S}_n(\bs{\theta})$.

\section{Application to weighted cross entropy losses}\label{sec:ce}

In the following, we show that the well-known (weighted) Cross Entropy (wCE)  loss \cite{Aurelio} can be included in our framework as a particular wSOL. To observe this, let us consider the following admissible weight function
\begin{equation*}
\begin{split}
        & w_{CE}(y_i,\hat{y}_{\bs{\theta}}(\bs{x}_i))=   -\omega_0(1-y_i)\frac{\log(1-\hat{y}_{\bs{\theta}}(\bs{x}_i))}{\hat{y}_{\bs{\theta}}(\bs{x}_i)}-\omega_1y_i\frac{\log(\hat{y}_{\bs{\theta}}(\bs{x}_i))}{1-\hat{y}_{\bs{\theta}}(\bs{x}_i)},
\end{split}
\end{equation*}
where $\omega_0,\omega_1>0$ are weight parameters. We remark that $w_{CE}$ is well-defined as the prediction is in $(0,1)$. Moreover, here we choose the uniform pdf on $[0,1]$ for the threshold $\tau$, that is, $f(\xi)\equiv1$. Therefore, by setting $w=w_{CE}$ and referring to \eqref{eq:pesos}, we compute
\begin{equation*}
    \begin{split}
    W_{\textrm{P}}&=
     -\int_{0}^{1}{\omega_0\frac{\log(1-\hat{y}_{\bs{\theta}}(\bs{x}_i))}{\hat{y}_{\bs{\theta}}(\bs{x}_i)}\mathbbm{1}_{\{\hat{y}_{\bs{\theta}}(\bs{x}_i)>\xi\}}\mathrm{d}\xi}\\
     &= -\omega_0\frac{\log(1-\hat{y}_{\bs{\theta}}(\bs{x}_i))}{\hat{y}_{\bs{\theta}}(\bs{x}_i)}\int_{0}^{1}{\mathbbm{1}_{\{\hat{y}_{\bs{\theta}}(\bs{x}_i)>\xi\}}\mathrm{d}\xi}\\
     &=  -\omega_0\frac{\log(1-\hat{y}_{\bs{\theta}}(\bs{x}_i))}{\hat{y}_{\bs{\theta}}(\bs{x}_i)}\hat{y}_{\bs{\theta}}(\bs{x}_i)\\
     &=  -\omega_0\log(1-\hat{y}_{\bs{\theta}}(\bs{x}_i)).
    \end{split}
\end{equation*}
Similarly, we find
\begin{equation*}
    W_{\textrm{N}}= -\omega_1\log(\hat{y}_{\bs{\theta}}(\bs{x}_i)),
\end{equation*}
and therefore
\begin{equation*}
\begin{split}
    & \mathbb{E}_{\tau}[\mathrm{wFP}(\tau)] = -\omega_0\sum_{i=1}^n{(1-y_i)\log(1-\hat{y}_{\bs{\theta}}(\bs{x}_i))},\\
    & \mathbb{E}_{\tau}[\mathrm{wFN}(\tau)] = -\omega_1\sum_{i=1}^n{y_i\log(\hat{y}_{\bs{\theta}}(\bs{x}_i))}.\\    
\end{split}
\end{equation*}
Finally, taking again the score $s^{CE}=s^{cost}$ (see \eqref{eq:scost}), here we get the wCE
{\small{\begin{equation*}
    \ell_{s_w^{CE}}(\mathcal{S}_n(\bs{\theta}))=  -\sum_{i=1}^n{\omega_0(1-y_i)\log(1-\hat{y}_{\bs{\theta}}(\bs{x}_i))+\omega_1y_i\log(\hat{y}_{\bs{\theta}}(\bs{x}_i))}.
\end{equation*}}}
Note that we can recover the classical binary CE by setting $\omega_0=\omega_1=0$.

\section{Applications to value-weighted scores}\label{sec:valuew}
\subsection{On the severity of false and missed alarms}

\begin{figure}
    \centering
    \includegraphics[width=0.7\columnwidth]{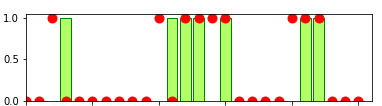}\\
    \includegraphics[width=0.7\columnwidth]{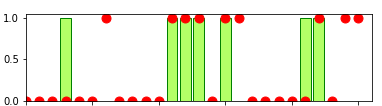}  
    \caption{Two different binary predictions with the same confusion matrix:  $\mathrm{TN}=15$, $\mathrm{TP}=5$, $\mathrm{FP}=4$ and $\mathrm{FN}=2$. The green bars are the binary true labels and the red dots are the binary predictions.}
    \label{fig:toy_ex}
\end{figure}

In classification problems with chronological data, the distribution of predictions along time with respect to the actual occurrences of events is not taken into account when the classical confusion matrix is computed. In Figure \ref{fig:toy_ex} we show two predictions which have the same CM, and therefore the scores, which are also referred to as \textit{skill scores} in forecasting applications, have the same value. However, the two predictions are different
from a forecasting value viewpoint. Indeed, the prediction in the first panel may be preferred since the missed events are anticipated by positive predictions, also called \textit{alarms}, whereas in the second panel the first event is completely missed, for example. The aim of value-weighted skill scores, which we detail in the next subsection, is then to differentiate the severity of errors by taking into account the sequential order. In such a way, the prediction in the first panel of Figure \ref{fig:toy_ex} will be associated to a higher value-weighted skill score than the one in second panel, since false positives and false negatives in the first prediction are less \textit{relevant} than the ones in the second.

\subsection{Value-weighted scores}

We therefore address the problem outlined in the previous subsection by defining a weight function that takes into account the \textit{value} of the error with respect to the chronological sequence of data. First, letting $T\in\mathbb{N}$, in this setting we consider the vectors (c.f. \eqref{eq:passato_futuro_pre})
\begin{equation}\label{eq:passato_futuro}
\begin{split}
    &\bs{y}^+_i= (y_{i+1},\dots,y_{i+T}),\quad \bs{\mathbbm{1}}^-_i= \big(\mathbbm{1}_{\{\hat{y}_{\bs{\theta}}(\bs{x}_{i-1})>\tau\}},\dots,\mathbbm{1}_{\{\hat{y}_{\bs{\theta}}(\bs{x}_{i-T})>\tau\}}\big),
\end{split}
\end{equation}
which provide information on future labels and past predictions, ordered with respect to the time distance from the \textit{present} referring index $i$. Exploiting the theoretical framework proposed in the previous sections, and inspired by \cite{guastavino2021}, we then restrict to the following structure for a value-weight function
\begin{equation*}
    w_{value}(y_i,\bs{y}^+_i,\bs{\mathbbm{1}}^-_i)=1-g(y_i\bs{\mathbbm{1}}^-_i+(1-y_i)\bs{y}^+_i).
\end{equation*}
where we assume $0\le g(\cdot)<1$. Let us elaborate on the consequences of this formulation. Note that $0<w_{value}(y_i,\bs{y}^+_i,\bs{\mathbbm{1}}^-_i)\le 1$, thus we are going to reward \textit{good} errors in place of penalizing \textit{bad} errors. Furthermore, we get
\begin{itemize}
    \item 
    $w_{value}(y_i,\bs{y}^+_i,\bs{\mathbbm{1}}^-_i)=1-g(\bs{y}^+_i)$ if $y_i=0$ (false positive) or
    \item
    $w_{value}(y_i,\bs{y}^+_i,\bs{\mathbbm{1}}^-_i)=1-g(\bs{\mathbbm{1}}^-_i)$ if $y_i=1$ (false negative).
\end{itemize}
This property agrees with the fact that a false alarm can be rewarded if it anticipates positive events occurring in the near \textit{future}, while, on the other hand, a missed alarm can be rewarded if alarms have been raised in the near \textit{past}.

In order to provide concrete examples and calculations for this setting, we focus on two particular formulations for the function $g$. In both cases, we make use of a positive weight vector $\bs{\omega}=(\omega_1,\dots,\omega_T)$ to act on $\bs{y}^+_i,\bs{\mathbbm{1}}^-_i$. We assume $\bs{\omega}$ to be non-increasing with respect to the indexing, i.e., $\omega_i\ge\omega_{i+1}$, $1\le i \le T-1$. Indeed, we expect the outcomes that are closer to the referring index $i$ to be more important than the ones that approach the extremum $i+T$ or $i-T$, as they are more distant in time. We focus on the following cases.
\begin{enumerate}
    \item 
    Case $g_{prod}(\bs{z})=\bs{\omega}\cdot\bs{z}$. With this structure, all the positive labels in the temporal window $[i+1,i+T]$ and alarms in $[i-1,i-T]$ influence the entrances of the matrix, in a way that is ruled by the weight vector $\bs{\omega}$.
    \item
    Case $g_{max}(\bs{z})=\max(\bs{\omega}\odot\bs{z})$, where $\odot$ is the pointwise product between two vectors. In this situation, being $\bs{\omega}$ non-increasing with respect to the indexing, only the closest positive label and alarm play a role in determining the value of wFP and wFN, respectively.
\end{enumerate}
\begin{remark}
    Let $w=w_{value}$. Since $0< w(\cdot)\le 1$, we have $\textrm{wFN}\le \textrm{FN}$ and $\textrm{wFP}\le \textrm{FP}$. Moreover, the scores are non-increasing with respect to false positives and negatives and thus $s\le s_w$ for any batch $\mathcal{S}_n(\bs{\theta})$.
\end{remark}

\subsection{wSOLs for value-weighted scores}

The structure of wSOLs in the case $w=w_{value}$ is in fact fully characterized by the formulations of the expected values of wFP and wFN, on which we focus. We obtain
\begin{equation*}
         \mathbb{E}_{\tau}[\mathrm{wFP}(\tau)] = \sum_{i=1}^n{(1-g(\bs{y}^+_i))(1-y_i)F(\hat{y}_{\bs{\theta}}(\bs{x}_i))},
\end{equation*}
because the weight does not depend on the threshold $\tau$. On the other hand, we have
\begin{equation*}
\begin{split}
 \mathbb{E}_{\tau}[\mathrm{wFN}(\tau)]&= \sum_{i=1}^n{y_i \mathbb{E}_{\tau}[(1-g(\bs{\mathbbm{1}}^-_i(\tau)))\mathbbm{1}_{\{\hat{y}_{\bs{\theta}}(\bs{x}_i)<\tau\}}]}\\
    &=\sum_{i=1}^n{y_i \mathbb{E}_{\tau}[\mathbbm{1}_{\{\hat{y}_{\bs{\theta}}(\bs{x}_i)<\tau\}}-g(\bs{\mathbbm{1}}^-_i(\tau))\mathbbm{1}_{\{\hat{y}_{\bs{\theta}}(\bs{x}_i)<\tau\}}]}\\
     &=\sum_{i=1}^n{y_i \big(1-F(\hat{y}_{\bs{\theta}}(\bs{x}_i))-\mathbb{E}_{\tau}[g(\bs{\mathbbm{1}}^-_i(\tau))\mathbbm{1}_{\{\hat{y}_{\bs{\theta}}(\bs{x}_i)<\tau\}}]\big)}.
    \end{split}
\end{equation*}
In the following, we will discuss separately the form of $\mathbb{E}_{\tau}[\mathrm{wFN}(\tau)]$ depending on the weight formulations outlined in the previous section. We may adopt the abridged notation $\hat{y}_{i}=\hat{y}_{\bs{\theta}}(\bs{x}_i)$.

\subsection{Case $g=g_{prod}$}
In this case, to satisfy the requirements on the weight function, we need $\sum_{i=1}^T{\omega_i}< 1$, i.e., $\lVert \bs{\omega}\lVert_1< 1$. We immediately get
\begin{equation*} 
\mathbb{E}_{\tau}[\mathrm{wFP}(\tau)] = \sum_{i=1}^n{(1-\bs{\omega}\cdot\bs{y}^+_i)(1-y_i)F(\hat{y}_i)}.
\end{equation*}
As far as the false negatives are concerned, we get the following.
\begin{theorem}\label{teor:prod_thm}
    Let $\hat{y}_{\bs{\theta}}(\bs{x}_{i-1}),\dots,\hat{y}_{\bs{\theta}}(\bs{x}_{i-T})$ be the $T$ predictions in chronological order before $\hat{y}_{\bs{\theta}}(\bs{x}_{i})$, and let $g=g_{prod}$. By defining for each $i=1,\dots,n$
    \begin{equation}\label{eq:diff}
    \begin{split}
        D_{i}(j)&=F(\hat{y}_{i-j})-F(\textrm{min}\{\hat{y}_i,\hat{y}_{i-j}\})\\
        &= \max\big\{F(\hat{y}_{i-j})-F(\hat{y}_i),0\big\},
    \end{split}
    \end{equation}
    we have
    \begin{equation*}
            \mathbb{E}_{\tau}[\mathrm{wFN}(\tau)] = \sum_{i=1}^n{y_i\bigg(1-F(\hat{y}_i)-\sum_{j=1}^T\omega_jD_i(j)\bigg)}.
    \end{equation*}
\end{theorem}
\begin{proof}
    We need to calculate the integral
\begin{equation*}
\begin{split}
        & \mathbb{E}_{\tau}[w\big(\bs{\mathbbm{1}}^-_i(\tau)\big)\big(1-\mathbbm{1}_{\{\hat{y}_i>\xi\}}\big)]=\\
        & \int_{a}^{b}{(1-\bs{\omega}\cdot (\mathbbm{1}_{\{\hat{y}_{i-1}>\xi\}},\dots,\mathbbm{1}_{\{\hat{y}_{i-T}>\xi\}}))(1-\mathbbm{1}_{\{\hat{y}_i>\xi\}})f(\xi)\mathrm{d}\xi}=\\
        & 1-F(\hat{y}_i)-\int_{a}^{b}{(\bs{\omega}\cdot (\mathbbm{1}_{\{\hat{y}_{i-1}>\xi\}},\dots,\mathbbm{1}_{\{\hat{y}_{i-T}>\xi\}}))(1-\mathbbm{1}_{\{\hat{y}_i>\xi\}})f(\xi)\mathrm{d}\xi}=\\
        & 1-F(\hat{y}_i)-\sum_{j=1}^T\omega_jF(\hat{y}_{i-j})+\int_{a}^{b}{\mathbbm{1}_{\{\hat{y}_i>\xi\}}(\bs{\omega}\cdot (\mathbbm{1}_{\{\hat{y}_{i-1}>\xi\}},\dots,\mathbbm{1}_{\{\hat{y}_{i-T}>\xi\}}))f(\xi)\mathrm{d}\xi}=\\
        & 1-F(\hat{y}_i)-\sum_{j=1}^T\omega_jF(\hat{y}_{i-j})+\sum_{j=1}^T \omega_jF(\textrm{min}\{\hat{y}_{i},\hat{y}_{i-j}\})=\\
        & 1-F(\hat{y}_i)-\sum_{j=1}^T\omega_j\big(F(\hat{y}_{i-j})-F(\textrm{min}\{\hat{y}_{i},\hat{y}_{i-j}\})\big).
\end{split}
\end{equation*}
Therefore
\begin{equation*}
\mathbb{E}_{\tau}[\mathrm{wFN}(\tau)] = \sum_{i=1}^n{y_i\bigg(1-F(\hat{y}_i)-\sum_{j=1}^T\omega_j\big(F(\hat{y}_{i-j})-F(\textrm{min}\{\hat{y}_{i},\hat{y}_{i-j}\})\big)\bigg)}.
\end{equation*}
\end{proof}
Let us comment on the achieved expression for $\mathbb{E}_{\tau}[\mathrm{wFN}(\tau)]$. In principle, in $g_{prod}(\bs{\mathbbm{1}}^-_i(\tau))=\bs{\omega}\cdot \bs{\mathbbm{1}}^-_i(\tau)$ the predictions that play a role are the ones that are larger than the fixed threshold value. When applying the expected value, we lose the possibility of deciding which predictions are positive in terms of a threshold. Looking at \eqref{eq:diff}, we observe that the prediction $\hat{y}_i$ substitutes $\tau$ in establishing if past predictions are to be considered positive or not. Indeed, only past predictions whose value is larger than $\hat{y}_i$ gives a non-zero contribute in the expression of $\mathbb{E}_{\tau}[\mathrm{wFN}(\tau)]$. Finally, note that if $\bs{\omega}\equiv \textbf{0}$ then $w\equiv 1$, and so $\mathbb{E}_{\tau}[\mathrm{wFP}(\tau)]=\mathbb{E}_{\tau}[\mathrm{FP}(\tau)]$ and $\mathbb{E}_{\tau}[\mathrm{wFN}(\tau)]=\mathbb{E}_{\tau}[\mathrm{FN}(\tau)]$, which is consistent with the original value-weighted score.


\subsection{Case $g=g_{max}$}\label{sec:sec_max}
In this case, we require that $\max_{i=1,\dots,T}{\omega_i}< 1$, i.e., $\lVert \bs{\omega}\lVert_{\infty}< 1$. First, we get
\begin{equation*}
    \mathbb{E}_{\tau}[\mathrm{wFP}(\tau)] = \sum_{i=1}^n{(1-\max(\bs{\omega}\odot\bs{y}^+_i))(1-y_i)F(\hat{y}_i)}.
\end{equation*}
To analyze what happens for the false negatives, we need to fix some preliminary definitions. We make the following assumption.
\begin{assumption}\label{ass:inclusiva}
    Let $\hat{y}_{i-j}$, $j=1,\dots,T$, be a prediction involved in the construction of the vector $\bs{\mathbbm{1}}^-_i$. We assume that $\hat{y}_{i-j}\in(a,b)$, that is, it belongs to the support of the pdf chosen for the threshold.
\end{assumption}
The requirement in Assumption \ref{ass:inclusiva} is mild, as many pdfs are supported in the whole interval $(0,1)$ (e.g., the uniform one). Moreover, this requirement significantly simplifies the following presentations, and analogous results can be achieved in the more general case.

We proceed by introducing some necessary ingredients.
\begin{definition}\label{def:linked_interval}
Let $\hat{y}_{i-j}$, $j=1,\dots,T$, be a prediction involved in the construction of the vector $\bs{\mathbbm{1}}^-_i$. We denote as \textit{power interval} the interval $I_j\subseteq [a,b]$ constructed as follows:
$\xi \in I_j$ if and only if
\begin{enumerate}
    \item 
    $\hat{y}_{i-j}>\xi$ and
    \item
    $\hat{y}_{i-s}\le\xi$ for every $s<j$, $s\in\{1,\dots,T\}$.
\end{enumerate}    
\end{definition}
We observe that $I_j$ might be empty for some values of $j$. Moreover, considering Assumption \ref{ass:inclusiva}, it always takes the following forms:
\begin{enumerate}
    \item
    $I_1=[a,\hat{y}_{i-1})$ and is never empty;
    \item
    For $j>1$, $I_j=[\hat{y}_{i-k(j)},\hat{y}_{i-j})$ for some index $k(j)\in\{1,\dots,T\}$, $k(j)<j$.
\end{enumerate}
This fact motivates the following.
\begin{definition}\label{def:actual}
    For $j=2,\dots,N$, we will denote $\hat{y}_{i-k(j)}$ as the \textit{actual precursor} of $\hat{y}_{i-j}$. We include the case $j=1$ by saying that $\hat{y}_{i-k(1)}=a$.
\end{definition}
Let us present a couple of examples to clarify the setting.
\paragraph{Examples.} Let $T=4$.
\begin{itemize}
    \item 
    If $\hat{y}_{i-1}=0.5$, $\hat{y}_{i-2}=0.6$, $\hat{y}_{i-3}=0.1$ and $\hat{y}_{i-4}=0.8$, then $I_1=[a,\hat{y}_{i-1})$, $I_2=[\hat{y}_{i-1},\hat{y}_{i-2})$, $I_3=\emptyset$ and $I_4=[\hat{y}_{i-2},\hat{y}_{i-4})$. Therefore, $\hat{y}_{i-1}$ is the actual precursor of $\hat{y}_{i-2}$, which is in turn the actual precursor of $\hat{y}_{i-4}$.
    \item 
    If $\hat{y}_{i-1}=0.7$, $\hat{y}_{i-2}=0.2$, $\hat{y}_{i-3}=0.9$ and $\hat{y}_{i-4}=0.3$, then $I_1=[a,\hat{y}_{i-1})$, $I_2=\emptyset$, $I_3=[\hat{y}_{i-1},\hat{y}_{i-3})$ and $I_4=\emptyset$. Therefore, $\hat{y}_{i-1}$ is the actual precursor of $\hat{y}_{i-3}$.
    \end{itemize}
    The next step is to introduce the following subset. We define $\{\hat{y}_{i-t_1},\dots,\hat{y}_{i-t_S}\}\subseteq\{\hat{y}_{i-1},\dots,\hat{y}_{i-T}\}$, $S\le T$, to be such that
    \begin{enumerate}
    \item
    $t_1=1$.
    \item 
    For every $j=2,\dots,S$, $I_{t_j}\neq\emptyset$, $\hat{y}_{i-t_{j-1}}<\hat{y}_{i-t_{j}}$ and $\hat{y}_{i-t_{j-1}}$ is the actual precursor of $\hat{y}_{i-t_{j}}$. In this direction, we also define $\hat{y}_{i-t_{0}}=a$.
    \end{enumerate}
    We are now ready to express the expected value of the false negatives entrance of the weighted CM.
\begin{theorem}\label{teor:thm_max}
    Let $\hat{y}_{i-1},\dots,\hat{y}_{i-T}$ be the $T$ predictions in chronological order before $\hat{y}_i$, and let $g=g_{max}$. We have
    {\small
    \begin{equation*}
            \mathbb{E}_{\tau}[\mathrm{wFN}(\tau)] = \sum_{i=1}^n{y_i\bigg(1-F(\hat{y}_i)-\sum_{j=1}^S(\omega_{t_j}-\omega_{t_{j+1}})D_i(t_j)\bigg)}.
    \end{equation*}}
    where $D_i(\cdot)$ was defined in Theorem \ref{teor:prod_thm} and $\omega_{S+1}=0$.
\end{theorem}
\begin{proof}
    We compute
\begin{equation*}
\begin{split}
            & \mathbb{E}_{\tau}[w\big(\bs{\mathbbm{1}}^-_i(\tau)\big)\big(1-\mathbbm{1}_{\{\hat{y}_i>\tau\}}\big)]=\\
            & \int_{a}^{b}{(1-\max(\bs{\omega}\odot (\mathbbm{1}_{\{\hat{y}_{i-1}>\xi\}},\dots,\mathbbm{1}_{\{\hat{y}_{i-T}>\xi\}})))(1-\mathbbm{1}_{\{\hat{y}_i>\xi\}})f(\xi)\mathrm{d}\xi}=\\
            & 1-F(\hat{y}_i)-\int_{a}^{b}{\max(\bs{\omega}\odot (\mathbbm{1}_{\{\hat{y}_{i-1}>\xi\}},\dots,\mathbbm{1}_{\{\hat{y}_{i-T}>\xi\}}))(1-\mathbbm{1}_{\{\hat{y}_i>\xi\}})f(\xi)\mathrm{d}\xi}.
\end{split}
\end{equation*}
Assume that $j\in\{1,\dots,T\}$ is the minimum index that realizes $\mathbbm{1}_{\{\hat{y}_{i-j}>\xi\}}=1$. Then 
$$
\max(\bs{\omega}\odot (\mathbbm{1}_{\{\hat{y}_{i-1}>\xi\}},\dots,\mathbbm{1}_{\{\hat{y}_{i-T}>\xi\}}))=\omega_{j}\mathbbm{1}_{\{\hat{y}_{i-j}>\xi\}}.
$$
To calculate the integral, we will make use of Definitions \ref{def:linked_interval} and \ref{def:actual}. We have
\begin{equation*}
        \mathbb{E}_{\tau}[w\big(\bs{\mathbbm{1}}^-_i(\tau)\big)\big(1-\mathbbm{1}_{\{\hat{y}_i>\tau\}}\big)]=1-F(\hat{y}_i)-\sum_{j=1}^T \int_{I_j}{\omega_{j}\mathbbm{1}_{\{\hat{y}_{i-j}>\xi\}}(1-\mathbbm{1}_{\{\hat{y}_i>\xi\}})f(\xi)\mathrm{d}\xi}
\end{equation*}
Let us focus on the computation of
\begin{equation*}
\begin{split}
    \sum_{j=1}^T \int_{I_j}{\omega_{j}\mathbbm{1}_{\{\hat{y}_{i-j}>\xi\}}(1-\mathbbm{1}_{\{\hat{y}_i>\xi\}})f(\xi)\mathrm{d}\xi}&=\int_a^{\hat{y}_{i-1}}{\omega_{1}\mathbbm{1}_{\{\hat{y}_{i-1}>\xi\}}(1-\mathbbm{1}_{\{\hat{y}_i>\xi\}})f(\xi)\mathrm{d}\xi}+\\
    &+\sum_{j=2}^T \int_{\hat{y}_{i-k(j)}}^{\hat{y}_{i-j}}{\omega_{j}\mathbbm{1}_{\{\hat{y}_{i-j}>\xi\}}(1-\mathbbm{1}_{\{\hat{y}_i>\xi\}})f(\xi)\mathrm{d}\xi},
\end{split}
\end{equation*}
where we assume $k(j)=j$ in the case $I_j=\emptyset$. We have
\begin{equation*}
\begin{split}
        & \int_a^{\hat{y}_{i-1}}{\omega_{1}\mathbbm{1}_{\{\hat{y}_{i-1}>\xi\}}(1-\mathbbm{1}_{\{\hat{y}_i>\xi\}})f(\xi)\mathrm{d}\xi}=\\
        & \int_a^{\hat{y}_{i-1}}{\omega_{1}\mathbbm{1}_{\{\hat{y}_{i-1}>\xi\}}f(\xi)\mathrm{d}\xi}-\int_a^{\hat{y}_{i-1}}{\omega_{1}\mathbbm{1}_{\{\hat{y}_{i-1}>\xi\}}\mathbbm{1}_{\{\hat{y}_i>\xi\}}f(\xi)\mathrm{d}\xi}=\\
        &\omega_1F(\hat{y}_{i-1})-\omega_1F(\textrm{min}\{\hat{y}_{i-1},\hat{y}_{i}\})=\\
        &\omega_1\big(F(\hat{y}_{i-1})-F(\textrm{min}\{\hat{y}_{i-1},\hat{y}_{i}\})\big).
\end{split}
\end{equation*}
Similarly,
\begin{equation*}
\begin{split}
     & \int_{\hat{y}_{i-k(j)}}^{\hat{y}_{i-j}}{\omega_{j}\mathbbm{1}_{\{\hat{y}_{i-j}>\xi\}}(1-\mathbbm{1}_{\{\hat{y}_i>\xi\}})f(\xi)\mathrm{d}\xi}=\\
     & \int_{\hat{y}_{i-k(j)}}^{\hat{y}_{i-j}}{\omega_{j}\mathbbm{1}_{\{\hat{y}_{i-j}>\xi\}}f(\xi)\mathrm{d}\xi}-\int_{\hat{y}_{i-k(j)}}^{\hat{y}_{i-j}}{\omega_{j}\mathbbm{1}_{\{\hat{y}_{i-j}>\xi\}}\mathbbm{1}_{\{\hat{y}_i>\xi\}}f(\xi)\mathrm{d}\xi}=\\
      &\omega_j\big(F(\hat{y}_{i-j})-F(\hat{y}_{i-k(j)})\big)-\omega_j\big(F(\textrm{min}\{\hat{y}_{i-j},\hat{y}_{i}\})-F(\textrm{min}\{\hat{y}_{i-k(j)},\hat{y}_{i}\})\big)=\\
      & \omega_j\big(F(\hat{y}_{i-j})-F(\hat{y}_{i-k(j)})+F(\textrm{min}\{\hat{y}_{i-k(j)},\hat{y}_{i}\})-F(\textrm{min}\{\hat{y}_{i-j},\hat{y}_{i}\})\big).
\end{split}
\end{equation*}
By recalling that $\hat{y}_{i-k(1)}=a$, we can put together as
\begin{equation}\label{eq:max_final}
\begin{split}
        & \mathbb{E}_{\tau}[w\big(\bs{\mathbbm{1}}^-_i(\tau)\big)\big(1-\mathbbm{1}_{\{\hat{y}_i>\tau\}}\big)]=\\
        & 1-F(\hat{y}_i)-\sum_{j=1}^T\omega_j\big(F(\hat{y}_{i-j})-F(\hat{y}_{i-k(j)})+F(\textrm{min}\{\hat{y}_{i-k(j)},\hat{y}_{i}\})-F(\textrm{min}\{\hat{y}_{i-j},\hat{y}_{i}\})\big).    
\end{split}
\end{equation}
By using the subset of predictions introduced in Section \ref{sec:sec_max}, the result in \eqref{eq:max_final} can be rewritten as
\begin{equation*}
\begin{split}
        & \mathbb{E}_{\tau}[w\big(\bs{\mathbbm{1}}^-_i(\tau)\big)\big(1-\mathbbm{1}_{\{\hat{y}_i>\tau\}}\big)]=\\
        & 1-F(\hat{y}_i)-\sum_{j=1}^S\omega_{t_j}\big(F(\hat{y}_{i-t_j})-F(\hat{y}_{i-t_{j-1}})+F(\textrm{min}\{\hat{y}_{i-t_{j-1}},\hat{y}_{i}\})-F(\textrm{min}\{\hat{y}_{i-t_{j}},\hat{y}_{i}\})\big)=\\
        & 1-F(\hat{y}_i)-\sum_{j=1}^S(\omega_{t_{j}}-\omega_{t_{j+1}})\big(F(\hat{y}_{i-t_j})-F(\textrm{min}\{\hat{y}_{i-t_{j}},\hat{y}_{i}\})\big),
\end{split}
\end{equation*}
where we set $\omega_{S+1}=0$. Therefore
\begin{equation*}
\mathbb{E}_{\tau}[\mathrm{wFN}(\tau)] = \sum_{i=1}^n{y_i\bigg(1-F(\hat{y}_i)-\sum_{j=1}^S(\omega_{t_{j}}-\omega_{t_{j+1}})\big(F(\hat{y}_{i-t_j})-F(\textrm{min}\{\hat{y}_{i-t_{j}},\hat{y}_{i}\})\big)\bigg)}.
\end{equation*}
\end{proof}
The results obtained in Theorem \ref{teor:thm_max} shares a similar spirit with the one in Theorem \ref{teor:prod_thm}, but some differences ought to be highlighted. In $g_{max}(\bs{\mathbbm{1}}^-_i(\tau))=\max(\bs{\omega}\odot \bs{\mathbbm{1}}^-_i(\tau))$ the only prediction that plays a role is the closest alarm to the present index $i$. When applying the expected value, it is necessary for a past prediction to be larger than $\hat{y}_i$ to give a positive contribute in the expression of $\mathbb{E}_{\tau}[\mathrm{wFN}(\tau)]$. However, while this property is not only necessary but also sufficient in the $g_{prod}$ case, here we also need to refer to a \textit{hierarchy}, in which only the predictions that \textit{dominate} subsequent ones for some threshold values in the respective power interval are to be considered. This is observable in Definition \ref{def:linked_interval}, where indeed for a threshold $\xi\in I_j$ the value $\hat{y}_{i-j}$ acts as a temporary maximum with respect to the subsequent predictions $\hat{y}_{i-s}\le\xi$, $s<j$. 
Furthermore, the discrepancy between \textit{adjacent} elements of the weight vector $\bs{\omega}$ rules severely the influence of the past predictions. A noteworthy particular case is $\bs{\omega}\equiv \mathbf{1}$, where we observe that
\begin{equation*}
W_{\mathrm{N}}=1-F(\hat{y}_i)-\omega_{t_{j^{\star}}}\big(F(\hat{y}_{i-t_{j^{\star}}})-F(\textrm{min}\{\hat{y}_{i-t_{j^{\star}}},\hat{y}_{i}\})\big)
\end{equation*}
being $j^\star=\argmax_{j=1,\dots,S}{\hat{y}_{i-t_{j}}}$.

\section{Generalization to the multilabel framework}\label{sec:multilabel}

In this section, we discuss how our proposed theory can be extended to the multilabel classification setting. We recall that in the multilabel case each sample may belong to one or more classes, differently with respect to the classical multiclass framework where each element of the dataset is associated to one class only. The generalization is based on a one-versus-rest approach: letting $d\in\mathbb{N}$ be the number of classes, we can consider $d$ $2\times 2$ confusion matrices
$$
\mathrm{wCM}_1(\tau_1,\mathcal{S}_n(\bs{\theta})),\dots,\mathrm{wCM}_d(\tau_d,\mathcal{S}_n(\bs{\theta})),
$$
one for each class, where each $\mathrm{wCM}_i(\tau_i,\mathcal{S}_n(\bs{\theta}))$ describes the classification outcomes, achieved with respect to a certain threshold value $\tau_i$, in predicting class $i$ against the union of the remaining classes. Doing in this way, according to our theory, we can then consider the application of a classification metric $s$ to such expected confusion matrices, that is
$$
s\big(\mathbb{E}_{\tau_1}[\mathrm{wCM}_1(\tau_1,\mathcal{S}_n(\bs{\theta}))]\big),\dots,s\big(\mathbb{E}_{\tau_d}[\mathrm{wCM}_d(\tau_d,\mathcal{S}_n(\bs{\theta}))]\big),
$$
where $\tau_1,\dots,\tau_d$ can be treated as random variables in $[0,1]$ with possible different prior probability density functions. Finally, the scores can be gathered together by considering a global score
$$
S=\mu\big(s\big(\mathbb{E}_{\tau_1}[\mathrm{wCM}_1(\tau_1,\mathcal{S}_n(\bs{\theta}))]\big),\dots,s\big(\mathbb{E}_{\tau_d}[\mathrm{wCM}_d(\tau_d,\mathcal{S}_n(\bs{\theta}))]\big)\big),
$$
where $\mu:\mathbb{R}^d\longrightarrow\mathbb{R}$ is a function such as, e.g., the average of its arguments, the weighted average, the minimum. Then, we can define the loss function as outlined in Definition \ref{def:scoloss}. Note that in this extension to the multilabel setting the threshold parameter still plays a crucial role, differently with respect to the standard multiclass setting where the argmax function of the outputs of the network is taken into account.

\section{Conclusions}\label{sec:conclusions}

In this work, we first presented in details a theoretical framework that formalizes weighted classification metrics. Then, we showed how tailored losses can be designed for the optimization of such weighted scores in the training phase of the neural network. Finally, we proved the concreteness of the proposed setting by highlighting some of its particular instances that are well-known approaches that have been considered in literature. Therefore, the carried out analysis indicates that the constructed framework can be considered as a theoretical background for future developments of research lines and applications involving weighted classification scores and dedicated loss functions. 
\section*{Acknowledgements}
SG and FM were supported by the Programma Operativo Nazionale (PON) “Ricerca e Innovazione” 2014–2020. This research has been accomplished within GNCS-IN$\delta$AM.

\bibliographystyle{siam}

\bibliography{biblio}

\end{document}